# Artificial Intelligence for Dementia Research Methods Optimization

Magda Bucholc, PhD[a]*, Charlotte James, PhD[b]*, Ahmad Al Khleifat, PhD[c], AmanPreet Badhwar, PhD[d,e,f], Natasha Clarke, PhD[d], Amir Dehsarvi, PhD[g], Christopher R. Madan, PhD[h], Sarah J. Marzi, PhD[i,j], Cameron Shand, PhD[k], Brian M. Schilder, MPhil[i,j], Stefano Tamburin, MD PhD[l], Hanz M. Tantiangco, MSc[m], Ilianna Lourida, PhD[n], David J. Llewellyn, PhD[n,o^], Janice M. Ranson, PhD[n^]

[a] Cognitive Analytics Research Lab, School of Computing, Engineering & Intelligent Systems, Ulster University, Derry, UK

[b] NIHR Bristol Biomedical Research Centre, University Hospitals Bristol and Weston NHS Foundation Trust and University of Bristol, Bristol, UK

[c] Department of Basic and Clinical Neuroscience, Institute of Psychiatry, Psychology & Neuroscience, King's College London, London, United Kingdom

[d] Multiomics Investigation of Neurodegenerative Diseases (MIND) Lab, Centre de Recherche de l'Institut Universitaire de Gériatrie de Montréal, Montréal, Canada

[e] Institut de génie biomédical, Université de Montréal, Montréal, Canada

[f] Département de Pharmacologie et Physiologie, Université de Montréal, Montréal, Canada

[g] Aberdeen Biomedical Imaging Centre, School of Medicine, Medical Sciences, and Nutrition, University of Aberdeen, Aberdeen, UK

[h] School of Psychology, University of Nottingham, Nottingham, UK

[i] UK Dementia Research Institute, Imperial College London, London, UK

[j] Department of Brain Sciences, Imperial College London, London, UK

[k] Centre for Medical Image Computing, Department of Computer Science, University College London, London, UK

[l] Department of Neurosciences, Biomedicine and Movement Sciences, University of Verona, Verona, Italy

[m] Information School, University of Sheffield, Sheffield, UK

ⁿ University of Exeter Medical School, Exeter, UK

º The Alan Turing Institute, London, UK

**Corresponding author:** Magda Bucholc; School of Computing, Engineering & Intelligent Systems, Ulster University, United Kingdom, BT48 7JL. E-mail: [m.bucholc@ulster.ac.uk](mailto:m.bucholc@ulster.ac.uk)

* Joint first authors

^ Joint senior authors

## Abstract

**Introduction:** Machine learning (ML) has been extremely successful in identifying key features from high-dimensional datasets and executing complicated tasks with human expert levels of accuracy or greater.

**Methods:** We summarize and critically evaluate current applications of ML in dementia research and highlight directions for future research.

**Results:** We present an overview of ML algorithms most frequently used in dementia research and highlight future opportunities for the use of ML in clinical practice, experimental medicine, and clinical trials. We discuss issues of reproducibility, replicability and interpretability and how these impact the clinical applicability of dementia research. Finally, we give examples of how state-of-the-art methods, such as transfer learning, multi-task learning, and reinforcement learning, may be applied to overcome these issues and aid the translation of research to clinical practice in the future.

**Discussion:** ML-based models hold great promise to advance our understanding of the underlying causes and pathological mechanisms of dementia.

## 1. Introduction

Dementia is an age-related condition with increasing global prevalence and an annual global cost estimated at about US$1 trillion [1]. The timely detection of dementia is crucial in enabling effective disease management and providing optimal health care [2]. However, the complexity of underlying pathologies combined with considerable clinical heterogeneity present unique challenges to the development of effective treatments and early diagnostic tools for dementia. In recent years, developments in high-performance computing and machine learning (ML) algorithms have shown promise in improving dementia detection, monitoring, and management [3-5]. The use of ML methods has enabled analysis of large volumes of high-dimensional data, integration of various data sources (i.e., clinical, imaging, genetic), and identification of new disease associations and disease subtypes not previously discovered with traditional statistical approaches [6]. The application of ML algorithms has enabled the development of more flexible and scalable models that can advance our understanding of complex disease pathways with minimal human intervention. In this review, we provide a comprehensive summary of ML applications in dementia research, including the types of ML techniques being applied, the types of data they integrate, and their intended use, and discuss major opportunities and challenges of translating ML technologies from research to clinical practice.

## 2. Types of machine learning techniques used in dementia research

ML methods that have been applied in dementia research can broadly be divided into three classes: i) supervised learning; ii) unsupervised learning; and iii) semi-supervised learning

.

### 2.1    Supervised learning

The majority of ML approaches use supervised learning which is a subcategory of ML that uses labeled data to learn a target function that best maps input variables to an output [7,8]. Supervised learning is commonly separated into two types of problems, namely, classification and regression.

## 2.1.1  Classification approaches

Classification models seek to determine which of a set of pre-defined groups/categories an instance belongs to, given a set of labeled examples. In the context of dementia, classification approaches have been developed for disease detection, prognosis, and management [3,5]. Many of them have been derived from logistic regression [9-11], random forest (RF) [11,12], naïve Bayes [13,14], K-nearest neighbour (KNN) [14,15], decision trees [11], and support vector machines (SVM) [11,14-16].

Comparing these approaches, the most appropriate model is problem-dependent, and will be influenced by the type of data, the data collection procedure and its underlying distribution, the classification task itself (e.g. multi-class dementia status prediction), the training and optimization procedure that, in most cases, involves human intervention, and a host of other factors [15]. This is further complicated by various methods of model evaluation which, depending on the relative importance of different evaluation metrics, can lead to differing conclusions on which approach is optimal. Such evaluation metrics include: accuracy (the number of correct predictions made by a model in relation to the total number of predictions made); sensitivity (the ability to predict the condition when the condition is present); specificity (the ability to predict the absence of the condition when the condition is not present) and the classifier discriminant power (as estimated from the area under the Receiver Operating Characteristic [ROC] curve). In fact, different studies using different data sets have failed to generate one model that performed best in all applications [15]. While RF has shown advantages with non-linearly correlated data [12,17], SVM has demonstrated additional utility when there is a small number of samples and high number of features [11,18]. The results of a comparison of nine supervised models for predicting progression from mild cognitive impairment (MCI) to Alzheimer's Disease (AD) using cognitive reserve proxies showed that gradient-boosted decision trees achieved the best overall accuracy of 93% [19]. Gradient boosted decision trees, where each decision tree is trained to correct prediction errors of the previous tree, were also shown to outperform other classifiers when applied to miRNA expression (accuracy = 83%) [20] and memory clinic data (accuracy = 92%) [21].

Classification models in dementia have been built using many different amounts and types of input data [15,22]. Single-modality ML frameworks have been developed using cognitive and functional assessments (CFA) [22], magnetic resonance imaging (MRI) [23], positron emission tomography (PET) [24], cerebrospinal fluid (CSF) biomarkers [25], and genomic data [26], while more complex ML models used a combination of data inputs [27].

Evidence shows that classification approaches incorporating multiple data modalities generally lead to improvements in model performance [15, 28]. However, the individual contributions of data types to the overall performance of multi-modality ML frameworks are often not assessed, raising the question about the trade-off between performance and cost-effectiveness or efficiency of the proposed solutions. For example, Bucholc et al. [15] demonstrated that, although classifiers built using CFA and neuroimaging features performed consistently better than models based purely on CFA data, the accuracy of models incorporating both data types was only up to ~5% higher, with cost-effective CFA features being the most discriminative. While this demonstrates that incorporating neuroimaging data into a multi-modality framework can provide a performance increase, when it comes to translating research to clinical practice both high performance and clear interpretability for how this performance was achieved is required for adoption.

### 2.1.2   Regression approaches

Akin to classification, regression aims to learn the relationship between a continuous dependent variable and a number of independent variables. Several regression approaches have been applied in dementia studies [29-31]. In its simplest and most frequently used form, linear regression, a continuous dependent variable is regressed onto independent variables, a process during which coefficients in a linear model are estimated. Linear regression has been used nearly universally across different topics in dementia research, including to predict disease outcomes, estimate time to dementia, identify biomarkers, and subtype dementia phenotypes [32,33]. These studies span across varied modalities including clinical and phenotypic information [34], molecular measurements in peripheral tissues [35], neuroimaging [36], as well as multiple types of omic analyses in post-mortem brain [37].

Although linear regression is the most widely used regression model in dementia, the construction of a linear regression incorporating a large number of predictor variables often results in poor generalization performance. To ease this problem, different penalization functions have been proposed, each imposing different constraints. In the dementia context, penalized approaches have been shown to produce more stable results for correlated data and data where the number of predictors is much larger than the sample size [38]. Ridge regression performed especially well in the presence of high collinearity in linguistic data [39]. Lasso regression has shown some success in addressing high dimensional AD data, especially in the context of genetic risk detection [40], biomarker discovery [41], and analysis of neuroimaging-based endophenotypes [42]. Lastly, elastic net regression, which effectively combines lasso and ridge regression, has been shown to be a good compromise for variable selection and reduction of overfitting, while allowing for fast computational solutions and scaling to even more features than typical lasso regression. Elastic net regression has been used to study functional brain connectivity networks in AD brain [43], derive epigenetic biomarkers of Parkinson's disease [44], and classify AD and frontotemporal dementia (FTD) based on anatomical and functional imaging data [45].

Linear regression models are empirical models that only describe the observed data, without a true understanding of the underlying mechanism which generates the data. Non-linear regression approaches, where the function capturing the relationship between dependent and independent variables is more complex, are typically based on the underlying mechanisms which generate the data and, therefore, produce predictions that are often more reliable than linear models. In dementia research, linear regression models have been often ineffective in capturing nonlinear relationships between biomarkers (e.g. neuroimaging data) and cognitive measures, in particular when a small number of observations and a large number of features were used for model training [29]. On the other hand, nonparametric kernel-based methods, a non-linear approach, have achieved relatively robust estimates of the regression function [31]. A possible explanation is that non-linear models are more powerful and better capture the complex relationships between model input and output. As such, non-linear models have been successfully implemented not only to differentiate between stages of dementia severity, and improve risk prediction of AD, but also to identify potential descriptors for the decline of cognition using both single modality and multimodality data [15,31].

Supervised ML approaches are one of the most commonly implemented methods in dementia research [7,8]. However, both classification and regression often require a large amount of labeled data (especially when relationships are complex), making them less easily applied to e.g. research into rarer dementia subtypes, where large, labeled datasets are hard to obtain.

## 2.2    Unsupervised learning

Unlike supervised algorithms, unsupervised algorithms search for previously unknown patterns within unlabeled data sets. As such, they have particular utility in dementia studies where the labels (e.g. clinical diagnoses) are either unavailable [46] or uncertain [47].

Unsupervised learning comprises a wide variety of approaches, of which mixed-effects models [52], item response theory model [53], gaussian processes [54], kernel density estimation [55], and mixture models [56,57] are a few examples. Broadly, these models have been applied to identify disease trajectories or subtypes [52,57,58] or produce a progression risk score [53]. One group of approaches, disease progression models, aim to model trajectories of disease progression to predict decline and provide pathophysiological insights [59]. One such model, the event-based model [56], builds upon the hypothetical cascade model to obtain a sequence of events that describe one or more subtypes of disease progression [57,60].

The largest category of unsupervised learning methods is cluster analysis, where the aim is to find distinct groups within data, contingent on a suitable measure of similarity. Traditional clustering approaches, such as hierarchical clustering and density-based spatial clustering of applications with noise (DBSCAN), have been used to identify groups with, e.g., different rates of atrophy [48] or CSF biomarker profiles [49]. These methods do not, however, account for the temporal component of dementia [50] and are difficult to evaluate due to the lack of any ground truth [51].

As with supervised algorithms, the increasing richness of biological data across multiple modalities (imaging, fluid, clinical, genetics) provides further opportunities for unsupervised learning, with the potential to uncover complex relationships and elucidate the underlying pathophysiology [50,61]. This, and continuous methodological improvements in utilizing multi-modal data in a single model, increases the utility of unsupervised learning in extracting patterns from data. Recent applications include identifying a differential treatment response between data-driven subgroups [58] and the potential for reducing heterogeneity in clinical trials [53,61]. Nonetheless, the increased difficulty of model validation in unsupervised learning necessitates further work before clinical adoption becomes an option.

## 2.3    Semi-supervised learning

Semi-supervised learning falls between supervised and unsupervised ML, utilizing both labeled and unlabelled data. It is typically used in scenarios where there is a large amount of data available, yet only a small proportion of samples have been labeled.

Semi-supervised algorithms use the information from the unlabeled data points to improve the performance of a model trained on the small amount of labeled data [62]. Therefore, these approaches are most useful in applications where labeled data is limited, such as a lack of follow-up studies or the cost of acquiring the label (e.g. neuroimaging-based diagnoses). Different types of semi-supervised algorithms have been used for the classification of AD and MCI with datasets of different modalities, including brain imaging [23,63-65], among others [66].

There are many examples from dementia research that demonstrate the superiority of semi-supervised algorithms for diagnosis or prognosis, relative to supervised algorithms based upon more limited data. Batmanghelich et al. [64] presented a framework for dimensionality reduction that showed a semi-supervised algorithm outperformed supervised learning methods, for both classifier accuracy and area under the receiver operating characteristic curve (AUC). Filipovych and Davatzikos [65] confirmed that in some scenarios, e.g., in the absence of long-term follow-up evaluations, semi-supervised techniques may be more preferable to identify individuals with progressive disorders, such as those at risk of conversion from MCI to AD. The high performance of semi-supervised algorithms in predicting MCI to AD conversion was also demonstrated in Moradi et al. [23]. An et al. [63] developed a semi-supervised feature selection framework for diagnostic purposes using both imaging and genetic data, achieving superior performance (as defined by AUC) in different dementia prediction classification tasks compared to an SVM model using only labeled data. Furthermore, experimental results of semi-supervised distance metric learning with label propagation (SRF-LP) showed superior accuracy of SRF-LP compared with standard supervised learning algorithms, including RF, SVM, AdaBoost and  with an increase in the performance gap when the number of training samples was small [66].

Given the increasing amount of data available and the inherent uncertainty around labels (i.e., clinical diagnosis) in dementia research, semi-supervised learning provides the opportunity to combine prediction of clinically relevant features (e.g., global CDR score) with the utility of unsupervised learning, making the most of the available data. New techniques for semi-supervised learning are being developed, primarily centered around extending deep neural networks [62]. These methods remain underused in dementia research, though with researchers showing an increasing interest in deep learning algorithms, this is likely to change in the future.

## 2.4    Deep learning

Deep learning (DL) is a subset of ML inspired by the structure and information processing of biological neurons, which are organized into stacked layers to form a deep neural network. The main advantage of DL over traditional ML approaches is that the time-consuming steps of pre-processing and feature engineering of datasets are minimal and less critical since DL models are able to obtain new representations of data (e.g. combinations of biomarkers, DNA sequence motifs) via multiple non-linear transformations [67]. Success of DL in domains such as computer vision and natural language processing (NLP), an increasing availability of large datasets, and improvements in computational power have resulted in DL algorithms becoming more popular with dementia researchers in recent years. Most DL applications in dementia research have involved convolutional neural networks (CNNs) [68,69], autoencoders [70,71], and recurrent neural networks (RNNs) [72,73].

CNNs are typically applied to medical imaging data because they can capture multi-scale spatial information [74]. However, recent studies extended the use of CNNs by integrating multi-modal data [69]. For example, Spasov et al. [69] built a CNN model combining MRI, neuropsychological, demographic, and APOE data that achieved AUC of 0.93 when identifying subjects with MCI that converted to AD (versus stable MCI) over 3 years and AUC of 1 when differentiating between AD patients and control subjects. These results demonstrate that DL models incorporating multiple data modalities may become vital to fully utilize the wealth of information available for dementia research.

Autoencoders (AE) are a type of neural network that are used to project high dimensional data input into a latent low-dimensional code (encoder) by ignoring less informative features. They then reconstruct the input using the latent code. For example, an autoencoder was used to take, as input, multi-modal imaging markers (fluorodeoxyglucose PET (FDG PET), florbetapir PET, and structural MRI), to predict future decline to AD [70]. The output of the model was a novel trial enrichment criterion, known as the random denoising AE marker (rDAm), for identifying patients that are most likely to progress from MCI to AD [70]. Ithapu et al. [70] suggested that the use of the rDAm model could significantly improve our ability to design cost-effective AD trials, with smaller sample sizes and sufficient statistical power.

RNNs are typically used on data that is sequential or time-dependent in nature because their hidden state component (i.e. "memory cell") allows previous inputs to influence a given output [75]. For example, Alam et al. [72] applied a long short-term memory (LSTM) based RNN to predict onset of physical agitation episodes in patients with dementia, using motion sequences obtained from smartwatches. In this specific scenario, the LSTM-RNN model showed significantly better F1-scores (0.85) compared to traditional ML methods, such as KNN (0.69), SVM (0.67), and AdaBoost (0.71), highlighting the potential of using such models in sensing-based behavior inference.

There are several key challenges that limit the translation of DL methods into clinical practice. DL algorithms require a large amount of data: data processing, such as combining different biomarkers, is time-consuming and costly, and there is also lack of sufficient quality data, especially in multi-modality studies [76]. DL models are often referred to as "black-box" models due to non-linear feature transformations that emerge from multiple hidden layers, thereby reducing their interpretability when compared to other ML approaches. This lack of interpretability may lead to a lack of trust in the models, which can act as a barrier to implementation (see section 4.2 for more details).

## 3. Goals of the studies implementing machine learning approaches for dementia research

In dementia research, the goal of using ML approaches is to improve dementia prevention, diagnosis, treatment, and care. Due to the complex nature of dementia and the increasing size and complexity of datasets available, traditional statistical methods are often insufficient. The key areas of dementia research where ML can, or is, having a significant impact are clinical practice, experimental medicine, and clinical trials.

### 3.1   Machine learning in clinical practice

In clinical practice, ML has been applied to assessment of dementia risk, clinical diagnosis, prognosis, and care [3]. Of these, diagnostic studies have primarily focused on differentiating between stages of cognitive impairment or dementia subtypes [15,77]. Many have used neuroimaging data as input, mainly T1-weighted MRI and/or F-FDG PET. These data have been used to develop ML models for identifying the severity of cognitive impairment (e.g., classify normal controls vs MCI vs dementia/AD) and differentiating between dementia subtypes (e.g., AD vs FTD) [78]. More recently, studies have emerged that use a multi-modal framework to integrate heterogeneous data types such as demographic, neuropsychological, clinical, genetic, CSF or other omics data. Novel ML approaches, such as a kernel-based SVM classifier with a truncated singular value decomposition dimensionality reduction technique, have emerged to more effectively handle the heterogeneity of such data types and the additional complexity introduced by considering their multi-scale interactions [79]. Multi-modal data integration can be a very useful strategy for early detection of dementia status or susceptibility, more robustly identifying disease targets, and identifying causal links between different biomarkers, symptoms and clinical subtypes.

In the absence of established disease-modifying treatments for AD and other neurodegenerative diseases, the bulk of ML prognostic studies have focused on predicting the conversion from MCI to dementia using MRI [80], electroencephalography (EEG) [81], magnetoencephalography (MEG) [81], neuropsychological measures [82], genetic data [83] or combinations of modality types [84] . A recent systematic review of studies predicting MCI conversion to dementia included results of 234 experiments from 111 articles [85]. The authors found that, despite some methodological issues, incorporating domain-targeted cognitive measures and $^{18}$F-FDG PET data into a model results in its superior predictive performance over models built without these data types. Furthermore, the addition of other feature types does not significantly improve performance of ML models compared to using cognitive or FDG PET features alone. Similar observations have been made by Bucholc et al. [15]. They found that classifiers built using cognitive and functional assessments were the ones that performed consistently better than models based on other types of data, and that incorporating multi-modality features (e.g. cognitive and MRI or CSF data) into the predictive model provided only a small performance increase. In fact, when considering all studies, it appears that the improvement one gains by including other data types along with cognitive measures is often not significant [15,86,87]. This is somehow encouraging given the fact that cognitive measures can be easily collected in clinical routine, at a low cost. However, if specialist data such as neuroimaging and genetics becomes less expensive and more practical to collect in the future, classification approaches incorporating multiple data modalities, that can improve predictions even by a few percent, may become advantageous and their chance of implementation in clinical practice may increase.

Given the literature on ML approaches in dementia research, it is fair to say that a number of different algorithms have been shown to detect dementia and its prodromal phase with relatively high predictive accuracy, but their performance significantly varies when other performance metrics are considered [15]. In some cases, the high accuracy might have been arbitrarily increased by using a dataset with a large proportion of people without dementia [88]. From the clinical point of view, no single metric captures all the desirable properties of a model and therefore, it is important to have thorough understanding of, distinctions between, and uses and misuses of each of these metrics, especially in the context of clinical utility. For instance, Model 1 (e.g., used for automatic dementia screening) could identify ~10% of the population with probable dementia, generating a very high case load for clinicians to screen, review, and test. This would have high sensitivity (i.e., proportion of patients with actual dementia identified) but a low positive predictive value (PPV) (i.e., a proportion of those identified as having dementia in routinely collected data sets that are true dementia cases). Conversely, Model 2 could identify only 1% of the population as probable dementia, requiring clinical review. Even though this has lower sensitivity, it would be more efficient in having a higher PPV. Hence, it is essential to provide all the necessary information about a ML model, as well as the reference standard and the data used to develop it, to characterize a diagnostic process adequately.

Apart from distinctive differences in ML metrics for performance comparison, other issues need to be addressed before diagnostic, prognostic, and risk prediction algorithms can be routinely applied in settings such as memory clinics. These include reproducibility; model validation, and data leakage; generalizability of the model to data collected in different settings than the one used for model training and testing; and model interpretability [89]. Recently, a framework employing a transfer learning paradigm with ensemble learning algorithms (using multiple methods in tandem to make a consensus prediction) has been proposed for risk prediction of dementia at both population and individual levels [90]. In comparison to a baseline model, the target model, utilizing a parameter-transfer learning approach (training on larger, less task-specific datasets and then fine-tuning on smaller, more specific datasets) to update the decision boundaries of the baseline model, achieved better performance across all the performance metrics, including an increase in sensitivity of 19.1%, specificity of 2.7%, accuracy of 16.9%, and AUC of 11%. This shows the potential for transfer learning to overcome some of the big challenges of dementia research, such as regulatory challenges associated with data aggregation, management, privacy, and informed consent for the collection, use, and sharing of the data. In the future, some of the largest datasets used for dementia research (e.g. Alzheimer's Disease Neuroimaging Initiative (ADNI) or National Alzheimer's Coordinating Center (NACC)) could serve as source data for the development of machine learning models that smaller studies could use transfer learning to build upon.

Finally, the limitations of machine learning systems need to be clearly communicated to clinical end users, before a new generation of clinical decision support systems (CDSSs) designed to exploit the potentials of data-driven decision making are adopted and routinely used in clinical practice. Although a few studies have reported on the implementation of ML methodologies for determining the severity of dementia [15,91] and differential diagnosis of dementia in memory clinics [92], CDSSs for dementia diagnosis, prognosis, and management at the point of care are generally still underutilized.

## 3.2    Machine learning in experimental medicine

The theoretical number of unique small molecules is more than an order of magnitude greater than the number of atoms in the observable universe [93]. Therefore, optimizing the design of synthetic molecules for a desired therapeutic outcome is a computationally intractable task to perform by exhaustive brute force calculations. AI has the potential to much more efficiently search this complex space to design drugs that best approximate viable candidate molecules [94]. DL in particular has proven to be adept at these tasks as it is able to learn new representations of multi-modal data (e.g. neuroimaging, genomics, and clinical records) using nonlinear mappings [95]. As a consequence, there has been an explosion of applications to pharmacology and its related fields; chemoinformatics and structural biology [96]. Example use cases include: predicting 3D structure and function of small molecules from chemical formulas [97]; protein-protein or protein-drug interaction modeling; clinical outcome or biomarker prediction [98,99]; and personalized genome-drug interactions (i.e. pharmacogenomics). In addition to the design of new drugs, AI has also been applied to the repurposing of existing therapies developed for other kinds of treatment. Of particular relevance to dementia, a recent study by Dias et al [100] used the IBM Watson for Drug Discovery online tool, incorporating an NLP algorithm, to extract lists of gene-disease and drug-disease relationships from millions of published research articles related to the medical sciences. They then combined these relationships with gene co-expression data from human brain samples and created an extended knowledge network that revealed previously unknown relationships between different psychiatric and neurological disorders and hundreds of drugs. As a result, several drug candidates were identified to reposition as therapies for AD (n=25), Parkinson's Disease (n=1), and dementia (n=1) (see Supplementary Table 4 of original publication for details) [100]. Furthermore, potential associations between the pathological stage of AD and genes using a ML-based Drug Repurposing In AD (DRIAD) framework were evaluated in Rodriguez et al [101]. Here, DRIAD was applied to lists of genes that were differentially expressed after exposing neuronal cells to a test panel of 80 clinically approved drugs, generating a ranking of possible repurposing candidates that, after additional validation in relevant in vitro and in vivo AD model systems, could be evaluated in a clinical trial.

Outside the domain of drug discovery, AI has been applied in disease genomics, e.g., to predict the functional consequences of mutations in both protein-coding and non-coding genomic sequences [102,103]. These kinds of sequence-informed ML approaches hold several considerable advantages over explicit rule-based models, including the ability to then conduct *in silico* mutagenesis to probe the effects of all possible mutations [104]. These are powerful tools for functional impact prediction (e.g., gene expression, chromatin modification). However, when it comes to predicting disease status from genomic data, ML-based approaches have yet to show substantial performance increases over simpler additive methods like polygenic risk scores in autoimmune disease or AD [26]. This could be due to several factors including lack of sufficient sample sizes, the use of genotype arrays instead of whole-genome sequencing, the use of highly-processed input data (e.g.disease-associated variants identified through simple linear models), or the use of suboptimal ML architectures for this problem.

Another challenge is that sequence prediction models have not typically addressed the tissue- and cell-type-specific nature of mutational effects. Recent advances in single-cell transcriptomics, epigenomics and proteomics have permitted the accumulation of single-cell atlases in model organisms of dementia-related diseases, primary cells from living patients (e.g. blood) [105], postmortem samples (e.g. brain tissue) [106], and patient-derived induced pluripotent stem cells (iPSCs) [107]. Some dedicated databases have emerged for integrating and hosting AD-related single-cell datasets [108]. These datasets can subsequently be used to train DL models to learn latent representation of cell-type-specific responses to various genetic and chemical perturbations, as has been done in cancer cell lines previously [109], as well as natural genomic variation [110,111].Therefore, the evaluation of cell-type-specific effects of dementia-associated mutations is becoming increasingly feasible at scale.

## 3.3    Machine learning in clinical trials

The utilization of artificial intelligence and machine learning in clinical trials in Dementia is an area of research interest. A focal point of this research is the development of algorithms that possess the capability to identify and diagnose the disease in patients. This can enhance the precision of patient selection for clinical trials and aid in the monitoring of disease progression. Additionally, AI and ML techniques can be utilized to analyze vast amounts of data derived from clinical trials, such as electronic medical records and imaging data, to discern patterns and potential new treatment options [112].

Several studies have investigated the role of ML in clinical trial design [112,113]. Ezzati and Lipton [114] developed a ML framework incorporating the KNN algorithm, to identify individuals who were more likely to show cognitive decline during the follow-up and used this subgroup of participants for analysis of treatment effects. Their results indicated the ML model could provide ~17% and ~25% improvement in prediction of cognitive decline at 12 months and 24 months follow-up respectively, and hence, be effectively used to improve the power of clinical trials. Reith et al. [115] utilized baseline clinical information and CNN-extracted PET features to predict changes in quantitative biomarkers of brain pathology with gradient-boosted decision trees. The use of ML helped them identify a cohort with the fastest amyloid deposition, at a 2 to 4 times higher rate than random selection or other common selection methods used for patient recruitment. The potential of different ML approaches to assist in recruiting patients at risk of dementia has also been shown in other studies [116]. Hane et al. [116] showed that the incorporation of clinical notes into ML frameworks can aid model accuracy and can therefore be routinely used to identify individuals for interventions, such as disease management programs and screening for clinical trials. Another study applied a new ML approach, Subtype and Stage Inference (SuStaIn), to routinely acquired MRI scans from patients with dementia to identify different subtypes of dementia early on in the disease process [60]. The algorithm was able to determine three different subtypes of AD, which broadly matched those found in post-mortems of brain tissue.

It is increasingly recognized that Dementias are preceded by a pre-symptomatic or prodromal period of varying duration, during which the underlying disease process unfolds. This highlights opportunities to slow disease progression during different pre-symptomatic phases of the disease, when it is more likely that pathological changes can be delayed, arrested, or even reversed. Selecting study participants at high risk for Dementia or Mild Cognitive Impairment is therefore essential to design cost-effective prevention trials. The use of AI and ML in clinical trials for Dementia has the potential to increase the success, generalizability, and efficiency of these trials. These technologies can assist with patient recruitment and cohort composition, improve patient retention and protocol adherence, help process and manage large quantities of data from sources such as wearables and other smart devices, identify drug targets and candidate molecule generation, and discover subgroup effects. Ultimately, this can lead to the offering of new treatments to the right population at a faster pace [112,113].

Despite the large number of ML models that have been developed for dementia research, not many have been employed for patient–trial matching and recruitment before the start of a clinical trial or patient monitoring during the trial. There are still challenges to be addressed, including ensuring the safety and privacy of patient data, and addressing concerns regarding the interpretability of AI generated results. Further research is necessary to develop robust and validated AI and ML models that can be widely adopted within the clinical trial process for dementia.

## 4. Reproducibility, replicability, interpretability and clinical applicability issues in dementia research

### 4.1 Reproducibility and replicability

As we develop novel computational models to understand dementia, issues of reproducibility and replicability become increasingly important. While reproducibility involves obtaining the same results from the same model and data, replicability is based on applying the same model to independent datasets and observing generalizability of the findings.

Reproducibility is a long standing issue in scientific research, particularly experimental medicine, as results can vary due to both controllable (e.g. following identical protocols) and uncontrollable (e.g. system stochasticity) factors. The use of computational models therefore significantly improves reproducibility: an algorithm trained to perform a task will always give the same results when applied to the same data. However, in order for research to be fully reproducible, both data and code needs to be made available. A recent review of the use of ML for modeling progression to AD found that, while 75% of studies used publicly available data, only 7% shared their implementation code [117]. If guidelines such as the Turing Way and Findability Accessibility Interoperability and Reproducibility (FAIR) principles are routinely followed, the increasing use of AI in dementia research will allow results to be easily reproduced [118]

In contrast to reproducibility, replicability is much harder to achieve: an algorithm optimized to, for example, predict incident dementia in one cohort is not guaranteed to perform well in a completely different cohort. This is particularly true for algorithms, such as neural networks, that are susceptible to overfitting during training. One approach to improving replicability is to use data from multiple sources. There exist many large-scale studies that are easily accessible to dementia researchers, for example ADNI [119], Longitudinal Aging Study in India (LASI) [120], NACC [121], and BioBank [122]. In combination, these studies provide an opportunity to improve model generalization, however simply combining the data would be an enormous task due to differences in, for example, the reporting of variables or diagnostic criteria.

State-of-the-art ML methods such as transfer learning and multi-task learning, have the potential to improve replicability and generalizability of results. For example, transfer learning could be used to train an algorithm on data from more than one study, providing some variables in both data sets are the same, by finding a mapping between the data domains. Such an approach could help improve generalizability of the results and reduce overfitting to the cohort used for training, thus increasing the chance of replicability in further cohorts. Successful applications of transfer learning in dementia have been demonstrated in Danso et al. [90].

While transfer learning implies a sequentially shared representation, multi-task learning is related to a shared representation that is developed concurrently across different tasks. So far, only few studies implemented the multi-task ML approach in dementia context [123]. One example is the use of a Deep Feed Forward Neural Network (DFNN) approach based on multi-task learning (i.e. employing multiple loss functions) to simultaneously detect AD and determine its progression stage [123]. Khoei et al. [123] showed that the proposed model can accurately classify and predict AD and its stages using both binary and multi-class classification of AD (3 and 4 different class labels), over the time spans of 5 and 10 years.

## 4.2    Interpretability of machine learning models

Alongside reproducibility and replicability, the interpretability of ML algorithms is an important consideration for many applications, particularly in cases where the algorithm is making life-changing decisions. This is clearly the case for dementia research, and healthcare applications in general. In order for a clinical decision maker or patient to trust an algorithm, an understanding of how it makes its predictions is pertinent; lack of transparency represents a barrier to translation from research to clinical practice [124,125].

The interpretability of an algorithm can be divided into two classes: global interpretability and local interpretability. Global interpretability refers to how the components of an algorithm, such as features and weights, combine to make decisions. In contrast, local interpretability is associated with individual decisions; how did the algorithm make its decision about a specific sample or patient [126]. Global interpretability is hard to achieve due to the often-complex nature of ML algorithms. However, it is possible to interrogate algorithms at a modular level to understand how varying one feature or weight impacts the decisions being made.

One of the most common methods used in dementia research to interpret algorithmic decisions is (permutation) feature importance which does exactly this; it informs the user how much each feature contributes to the decisions being made [127]. One benefit of feature importance is that it is model agnostic and so can be used in combination with many ML algorithms. Model-specific approaches to interpretability are more often used alongside DL algorithms [128,129]. For example, Qiu et al. [130] constructed 'disease probability maps' to visualize how a fully convolutional network determined AD status from MRI images.

Common ML algorithms used in dementia research, such as decision trees, SVM and Bayesian networks, are inherently interpretable; they learn a set of rules or relationships between variables with a known structure that can be interrogated by the user. In contrast, 'black-box' models, including anything involving a neural network, are harder to interpret because the mathematical relationships between variables are learnt from the data and are inaccessible to the user. State of the art methods, such as local interpretable model-agnostic explanations (LIME) [131] and Shapley values [132], can be used to assess how a 'black-box' model came to a certain decision. However, while methods such as these exist and produce encouraging results, they are currently not routinely used in dementia research.

It may be questioned whether methods for interrogating 'black-box' models are necessary for dementia research. There are many examples in the literature where the performance of different ML algorithms has been compared. For the task of dementia diagnosis, interpretable algorithms often perform as well as, if not better than, 'black-box' models [133,134]. However, which algorithm performs best depends on the data, the application and how performance is measured [4,135]; interpretable algorithms can perform better in terms of predictive power [136] while 'black-box' models often achieve higher discriminative accuracy [135].

This boost in accuracy may make 'black-box' models seem attractive, but this is at the cost of interpretability. Recent reviews have found that while 'black-box' models are being used in dementia research, researchers still favor interpretable models [3,8]. In the future, if methods to interpret 'black-box' models are routinely implemented, and the use of ML becomes more common in clinical practice, this pattern may shift.

## 4.3    Clinical applicability issues

There are currently limited examples of ML models for dementia diagnosis, prognosis and management being successfully deployed into clinical practice. Issues associated with dataset collection, such as dataset shift (i.e. the scenario when the joint distribution of inputs and outputs differs between the training and test sets), omitted variable bias, unintended discriminatory bias, and repeated use of limited datasets, e.g., ADNI, result in impairing the ML model ability to generalize to new populations. Furthermore, technical issues related to difficulties in extracting patient data in a harmonized, machine-readable format; the lack of understanding of the mechanistic basis of model predictions; differences between clinical settings (e.g., including differences in equipment, coding definitions, and computer systems, such as electronic health records); and variations in local clinical practices further complicate the clinical adoption of ML solutions in dementia. Upon deployment, differences in the distribution of true patients and healthy individuals compared to that seen during training can also impact model performance. For example, a well-calibrated model will identify healthy individuals at a rate equal to that seen in the training data. If this is artificially high because proportionally fewer true patients volunteer for research, the model will predict a lower prevalence of disease [137].

The impact of these issues is demonstrated in the neuroimaging field, in which the mean squared error of a CNN trained to predict regional brain atrophy using ADNI data increased from 0.31 to 0.41 when tested on memory clinic data [138]. Performance improved when the training set included a wider range of scanner images and protocol types, suggesting a route to improved clinical application through increasing heterogeneity of training data [138]. Similar approaches, training on heterogeneous memory clinic data [139], or employing a transfer learning approach [90], have demonstrated improvements in generalizability. Better understanding of the genetic architecture of dementia has also been increased through transethnic genome wide association studies (GWAS), enhancing research previously limited to European-centric populations with more diverse populations [140].

Studies that trained models on cognitive data face similar issues. Promisingly, an AI-based iPad application to assess global cognition has demonstrated good performance detecting MCI and mild AD, with AUC of 0.81 and 0.88 respectively, using two linguistically and culturally distinct datasets [141]. Detecting dementia through speech using ML and NLP is a growing field, however, studies have largely focused on English speaking participants, limiting generalizability to other languages [142]. Recent studies have found that training on multilingual datasets improves performance (F-score = 0.85) compared to training on an English corpus alone (F-score = 0.80) [143].

Parallel issues that may limit clinical applicability are cost and availability of resources, e.g., relying on expensive imaging data. Multiomics approaches may increase this problem, and so the use of complementary and unique methods to find the most informative features will be key [144]. Models based on widely available and accessible data, such as wearables, may aid clinical translatability [145]. Collaboration between researchers and stakeholders will also improve clinical translation, such as working with clinicians to develop ML solutions for 'real world settings' [145]. A lack of trust in algorithmic decision making may also limit translation, compounded by opaque models with low interpretability; trust can be improved with transparent, explainable approaches [146], and reporting an algorithm's confidence in its decision [147]. In addition, performance metrics should aim to capture real clinical applicability and be easily understandable to clinicians. Given a lack of data on outcomes it is challenging to predict the impact of ML methods for dementia, but models should be rigorously tested prior to being deployed [138]. Clinical trials for ML interventions in dementia (similar to those designed for cancer research [148]) could evaluate the clinical utility of ML models given their inherent opacity and 'black-box' nature, as well as address the challenges related to generalizability and interpretability of ML models.

## 5. Discussion

### 5.1 Summary of existing methods and their limitations

It is clear that the use of AI and ML in dementia research is becoming increasingly common. Supervised learning algorithms are most frequently implemented: classification is commonly applied to predicting the risk of progression to dementia and differential diagnosis of dementia, while regression is used for detecting cognitive changes, evaluating dementia severity, estimating time to dementia, identifying risk factors [149] and dementia subtypes [38,150]. In situations where labeled data is not available or insufficient, unsupervised and semi-supervised approaches are being used with promising results [58,66]. Furthermore, the application of DL models for early detection and classification of dementia has gained considerable attention in recent years, mostly due to their improved performance over traditional ML approaches [69]. However, the successful implementation of ML models in clinical settings is currently a high-risk proposition due to their over-reliance on one data source; poor external validations; lack of or unsufficient understanding of the mechanistic basis of model predictions and clinical utility of different performance metrics; and potential bias caused by missing data or inappropriate use of methods for missing data imputations. Other sources of bias, including measurement bias, evaluation bias, sampling bias, and algorithm bias also impair the ability of ML models to generalize to new populations, and can lead to discrimination against specific groups or individuals. For example, bias rooted in unrepresentative datasets used for training and poor model calibration can lead to racial bias in model application, as noted in Gianattasio et al. [151]. In addition, the lack of generalizability of the findings of large datasets (e.g. ADNI, NACC, and similar, highly-selected clinical cohorts) to other populations, such as, community-based samples may critically limit their ability to produce ML models readily transferable to other settings [152]. The availability of data for ML algorithm development and validation in dementia is also a significant barrier to their adoption and routine use in clinical practice. The majority of ML models require large amounts of data for training and testing to ensure generalizability beyond the training set. The 'black-box' nature of many ML models, in particular DL models where input data can undergo complex transformations over many layers, means that they have no explicit declarative knowledge representation and hence, provide predictions without any accompanying justification. Other ML methods may be able to list dependencies between the target prediction output and the input features but those relationships are too complex to understand or verify. Finally, the objective comparison of models across dementia studies is challenging as obligatory performance metrics for certain model structures and use cases have not been defined. All these barriers to ML adoption require attention at each stage of model development, validation, and use, to enhance stakeholder trust (including regulatory agencies, researchers, clinicians, and industry partners) in the process and results.

## 5.2 Recommendations for Future Research

While the use of ML in dementia research presents many opportunities for the future, in order for algorithms to be translated to clinical practice, issues surrounding replicability, generalizability, interpretability, comparability, and trust in decisions need to be addressed.

*Replicability*

**Issue**: The majority of dementia studies implementing ML approaches assume that once the model performs well on the data for the specific domain or problem it is tasked with, the model can be successfully applied to new data. However, this assumption does not hold in many cases. Since the model is constructed for a specific problem/dataset, it has to be retrained and tuned for any new problem/dataset. This not only affects the time- and cost-effectiveness of the proposed solution but also may lead to the changes in model performance, often caused by discrepancies in data distribution or feature space.

**Recommendation**: Transfer learning, which allows re-using pre-trained models for domain-specific tasks, represents a time-saving alternative. The application of transfer learning frameworks in dementia research already shows some promising results, including the improved accuracy and prevention of overfitting when dealing with relatively small datasets [153].

*Generalizability*

**Issue**: Certain disease stages and patient characteristics are less likely to be well represented in datasets produced for clinical research. The lack of generalizability of clinical cohorts to other populations limits their ability to produce ML models that can adapt to other datasets. Furthermore, developing ML models using retrospective datasets does not always translate well to prospective applications.

The overwhelming majority of ML models used in dementia are purely associative, i.e., they focus on predicting outcomes based on a predefined set of variables. Since they are not constructed to understand causality between input data and output predictions, they cannot generalize well when changes in input data occur or when there are multiple possible causes of patient symptoms (e.g., differential diagnosis of dementia). The associative nature of ML algorithms for dementia places constraints on their performance and can lead to suboptimal diagnoses.

**Recommendation**: To ensure generalizability of ML models beyond the training data set, developers should use multiple datasets to test the stability of model performance. In addition, causal inference i.e., understanding how diagnosis is obtained and clearly defining the desired output, can be utilized to produce more robust and generalizable ML models [154]. Clinical trials for ML interventions in dementia can evaluate the clinical utility of ML models, including potential pitfalls in their practical deployment.

*Interpretability*

**Issue**: Interpretability and trust in algorithmic decisions are barriers for research being translated to a clinical setting.

**Recommendation**: Methods such as LIME and Shapley values can be used to improve interpretability and trust in models. These methods allow clinicians to interrogate the decision-making process of ML models, thus increasing transparency.

*Comparability*

**Issue**: Depending on the intended use case of a ML model, different metrics give different interpretations of the model performance.

**Recommendation**: Obligatory performance metrics for certain model types and use cases can be introduced to ensure that models with similar intended uses can be compared [155]. These metrics should be regularly re-assessed to test for potential model drift.

**Conclusions**

Improving generalizability and reproducibility of models will require the dementia data on which they are built to be widely available and accessible to other researchers, as well as representative of the population to which the model would be applied in a clinical setting. In the future, state-of-the-art methods such as transfer learning, causal modeling, reinforcement learning, and multi-task learning may be implemented. Methods such as these can not only be used in situations where labeled data is scarce, they can also be used to improve the generalizability of models, increasing validity across populations and, consequently, clinical applicability. If dementia researchers work with clinicians to overcome barriers to clinical translation, ML holds promise to change the landscape of dementia research and care in the future.

## Acknowledgments

With thanks to collaborators from the Deep Dementia Phenotyping (DEMON) Network State of the Science symposium participants (in alphabetical order): Peter Bagshaw, Robin Borchert, Magda Bucholc, James Duce, Charlotte James, David Llewellyn, Donald Lyall, Sarah Marzi, Danielle Newby, Neil Oxtoby, Janice Ranson, Tim Rittman, Nathan Skene, Eugene Tang, Michele Veldsman, Laura Winchester, Zhi Yao.

## Funding

This paper was the product of a DEMON Network state of the science symposium entitled "Harnessing Data Science and AI in Dementia Research" funded by Alzheimer's Research UK. JMR and DJL are supported by Alzheimer's Research UK and the Alan Turing Institute/Engineering and Physical Sciences Research Council (EP/N510129/1). DJL has also received funding from the National Institute for Health Research (NIHR) Applied Research Collaboration (ARC) South West Peninsula, National Health and Medical Research Council (NHMRC), JP Moulton Foundation, National Institute on Aging/National Institutes of Health (RF1AG055654), Alan Turing Institute/Engineering and Physical Sciences Research Council (EP/N510129/1). This work was additionally supported by Alzheimer's Research UK (MB, CJ), European Union INTERREG VA Programme (MB), Dr George Moore Endowment for Data

Science at Ulster University (MB), National Institute for Health and Care Research Bristol Biomedical Research Centre (CJ), Fonds de recherche du Québec Santé - Chercheur boursiers Junior 1 (AB), Canadian Consortium for Neurodegeneration in Aging and the Courtois Foundation (AB, NC), The Motor Neurone Disease Association Fellowship (Al Khleifat/Oct21/975-799) (AAK), ALS Association Milton Safenowitz Research Fellowship (grant number 22-PDF-609. doi: 10.52546/pc.gr.150909) (AAK), NIHR Maudsley Biomedical Research Centre (AAK), The Darby Rimmer Foundation (AAK), UKRI Future Leaders Fellowship (MR/S03546X/1) (CS), E-DADS project (EU JPND) (CS), EuroPOND project (EU Horizon 2020, no. 666992) (CS). SJM is funded by the Edmond and Lily Safra Early Career Fellowship Program and the UK Dementia Research Institute, which receives its funding from UK DRI Ltd, funded by the UK Medical Research Council, Alzheimer's Society and Alzheimer's Research UK.

**Conflicts of interest**

All authors declare no conflicts of interest.